\pdfoutput=1
\documentclass[11pt]{article}

\usepackage[final]{acl}

\usepackage{times}
\usepackage{latexsym}

\usepackage[T1]{fontenc}
\usepackage[utf8]{inputenc}

\usepackage{microtype}

\usepackage{inconsolata}

\usepackage{graphicx}
\usepackage{amsmath}
\usepackage[symbol]{footmisc}
\usepackage{csquotes}
\usepackage{epigraph}
\usepackage{listings}
\usepackage{hyperref}
\usepackage{booktabs}
\usepackage{multirow}

\usepackage{enumitem}

\usepackage{amssymb}% http://ctan.org/pkg/amssymb
\usepackage{pifont}% http://ctan.org/pkg/pifont

\usepackage{dblfloatfix}
\usepackage{amsthm}

\usepackage[most]{tcolorbox}
\usetikzlibrary{shadows.blur}

\makeatletter
\def\thm@space@setup{\thm@preskip=0pt
\thm@postskip=0pt}
\makeatother
\newtheoremstyle{newstyle}      
{5pt} %Aboveskip 
{} %Below skip
{\itshape} %Body font e.g.\mdseries,\bfseries,\scshape,\itshape
{} %Indent
{\bfseries} %Head font e.g.\bfseries,\scshape,\itshape
{.} %Punctuation afer theorem header
{ } %Space after theorem header
{} %Heading

\tcbset{
    mymsg/.style={
        enhanced,
        colframe=white, frame hidden,
        boxrule=0pt,
        arc=18pt,
        outer arc=18pt,
        boxsep=2pt, left=8pt, right=8pt, top=6pt,
        colback=#1!25,
        fuzzy shadow={1mm}{-0.7mm}{0mm}{0.1mm}{black!40!white},
        width=0.8\linewidth,
        before skip=-0.5mm, 
    },
    expertstyle/.style={mymsg=green, enlarge left by=10mm},
    llmstyle/.style={mymsg=blue, enlarge right by=10mm}
}

\theoremstyle{newstyle}
\newtheorem{theorem2}{Research Question}[section]



\title{Large Language Model Recall Uncertainty is Modulated by the Fan Effect}

\author{
  Jesse Roberts\\
  Tennessee Tech University \\
  \texttt{jtroberts@tntech.edu} 
   \\ \And
   Kyle Moore\\
   Vanderbilt University \\
   \texttt{kyle.a.moore@vanderbilt.edu} 
   \\ \AND
   Thao Pham \\
   Berea College \\
   \\ \And
   Oseremhen Ewaleifoh \\
  Vanderbilt University \\
   \\ \And
   Doug Fisher \\
   Vanderbilt University \\
}

\setlist{nosep}

\begin{document}
\maketitle
\begin{abstract}
% This paper evaluates whether large language models (LLMs) exhibit cognitive fan effects, similar to those discovered by Anderson in humans, after being pre-trained on human textual data. We conduct two sets of in-context recall experiments designed to elicit fan effects. Consistent with human results, we find that LLM recall uncertainty, measured via token probability, is influenced by the fan effect when attempting to reason about absent items. The experiments suggest the fan effect is consistent whether the fan value is induced in-context or in the pre-training data. Finally, these findings provide \textit{in-silico} evidence that fan effects and typicality are expressions of the same phenomena.
This paper evaluates whether large language models (LLMs) exhibit cognitive fan effects, similar to those discovered by Anderson in humans, after being pre-trained on human textual data. We conduct two sets of in-context recall experiments designed to elicit fan effects. Consistent with human results, we find that LLM recall uncertainty, measured via token probability, is influenced by the fan effect. Our results show that removing uncertainty disrupts the observed effect. The experiments suggest the fan effect is consistent whether the fan value is induced in-context or in the pre-training data. Finally, these findings provide \textit{in-silico} evidence that fan effects and typicality are expressions of the same phenomena.
\end{abstract}

\section{Introduction}

\footnotetext{https://github.com/JesseTNRoberts/Large-Language-Model-Recall-Uncertainty-is-Modulated-by-the-Fan-Effect}
    
Some subfields of AI are explicitly interested in understanding and mimicking the nature of human cognition (cognitive modeling, computational psychology, affective computing) but even more implicitly rely on models of human cognition (human-computer interaction, embodied robotics, collaborative robotics, AI assistive technology, computational game theory). A model that, through training, learned to implicitly exhibit human-like cognitive behaviors could be of tremendous value both to the explicit study of human cognition as an ethical test subject, and as a more faithful model of human behavior to those fields that seek to develop systems to work along side human counterparts. We believe that some large language models (LLM) may be excellent candidates for such a role. 

LLMs process information in a manner that is fundamentally different from humans. The matrix multiplications, maximum inner product search, and perceptron networks may have, at some level, been inspired by the biological neuronal system. But beyond the superficial, the systems bear no similarities. In spite of algorithmic and mechanistic dissimilarity, a growing body of work suggests that by merely training on human-language data, large language models learn to exhibit human-like cognitive behaviors as shown in Table \ref{tab:behav-rev}. 

In this paper, we survey the work applying cognitive science inspired evaluations to LLMs to analyze, understand, and catalog their relation to human cognition. We extend the existing work by providing the first investigation of human-like fan effects à la \citet{anderson1999fan} in LLMs. This effect is specifically interesting because it has a relation to the previously studied typicality effect, and it is understood to be an expression of human categorization uncertainty that has been precisely measured through response time delay.

Our results show that (1) some LLMs exhibit human-like fan effects based on the typicality of categorical items learned in pre-training; (2) some LLMs exhibit human-like fan effects based on the relative frequency of items in the model context; and (3) with uncertainty mitigated, the observed fan effect is disrupted. Of the models tested, Mistral \cite{jiang2023mistral} and SOLAR \cite{kim2023solar} exhibit noteworthy human-like fan effects, including nuanced differential fan effects previously observed in humans \cite{radvansky1999fan}. 

The results have two practical implications: LLMs learn to exhibit human-like uncertainty and that uncertainty may interfere with recall tasks. Our results additionally provide \textit{in-silico} evidence that the fan effect is a special case of typicality as is true in COBWEB models \cite{silber1989model}.

Understanding the cognitive behaviors acquired from language is essential to the successful application of LLMs in human-adjacent scenarios. Generally speaking, human-like cognitive effects may serve to smooth interactions between machine and human. Alternatively, a minority of discrepancies may serve to undermine the interactions.

\begin{table*}[h]
        \centering
        \resizebox{\linewidth}{!}{
        \begin{tabular}{l|c|c|c|c|c}
            Phenomena &
            Study by &
            \begin{tabular}[c]{@{}c@{}}Measure(s)\end{tabular} &
            \begin{tabular}[c]{@{}c@{}}Statistic \end{tabular} &
            \begin{tabular}[c]{@{}c@{}}Significance \end{tabular}&
            \begin{tabular}[c]{@{}c@{}}Systematic Perturbation \end{tabular}  \\
            \toprule
            \multirow{7}{*}{Theory of Mind}
                & \citet{bubeck2023sparks}       & qualitative             & ---                & ---   & ---     \\
                & \citet{kosinski2023theory}     & frequency             & ---                & ---      & ---   \\
                & \citet{sap2022neural}          & frequency               & ---                & ---   & ---    \\
                & \citet{ullman2023large}        & frequency               & ---                & ---       & ---     \\
                & \citet{trott2023large}         & token probs             & $\chi^2$ + $\beta$ & reported & ---   \\ 
                & \citet{ma2023towards}          & frequency               & ---                & ---    & --- \\
                & \citet{li2023theory}           & frequqncy               & ---                & ---    & --- \\
                
            \hline 
            \multirow{5}{*}{\begin{tabular}[c]{@{}l@{}}Logical  Reasoning\end{tabular}}
                & \citet{binz2023using}          & token probs             & $\chi^2$ + $t$ + $\beta$ & reported   & ---    \\
                & \citet{mccoy2019right}         & frequency               & ---       & ---  & ---    \\
                & \citet{lamprinidis2023llm}     & frequency               & ---       & ---  & ---  \\
                & \citet{yax2024studying}        & token probs             & $\chi^2$  & reported & --- \\
                & \citet{dasgupta2023language}   & frequency               & $\chi^2$ + $t$ & reported & --- \\
            \hline
            \multirow{3}{*}{\begin{tabular}[c]{@{}l@{}}Framing \& \\ Anchoring\end{tabular}}
                & \citet{binz2023using}          & token probs            & $\chi^2$ + $t$ + $\beta$ & reported    & ---    \\
                & \citet{jones2022capturing}     & frequency             & ---       & ---    & ---    \\
                & \citet{suri2023large}          & frequency             & ---        & reported   & ---   \\
            \hline
            \multirow{4}{*}{\begin{tabular}[c]{@{}l@{}}Decision-Making\end{tabular}}
                & \citet{binz2023using}          & token probs             & $\chi^2$ + $t$ + $\beta$ & reported   & ---   \\
                & \citet{jones2022capturing}     & frequency             & ---       & ---     & --- \\
                & \citet{coda2024cogbench}       & frequency             & $\beta$ & reported  & --- \\
                & \citet{hagendorff2023human}    & frequency             & $\chi^2$ & reported & --- \\
            \hline
            \multirow{2}{*}{\begin{tabular}[c]{@{}l@{}} Typicality \end{tabular}}
                & \citet{misra2021language}      & token probs             & $r$ + $\rho$  & reported   & ---    \\
    
                & \citet{roberts2024using}      & token probs             & $r$  & reported    & model   \\
            \hline
            \multirow{3}{*}{\begin{tabular}[c]{@{}l@{}} Priming \end{tabular}}
            
                & \citet{sinclair2022structural} & token probs             & --- & ---           & data  \\
                & \citet{roberts2024using}      & token probs              & $w$  & reported     & data + model \\
                & \citet{michaelov-etal-2023-structural} & token probs     & ---  & ---          & data   \\
            \hline
            \begin{tabular}[c]{@{}l@{}}Emotion  Induction\end{tabular}
                & \citet{coda2023inducing}       & frequency             & $r$ + $t$ + probit $\beta$   & reported     & ---   \\
        \end{tabular}}
        \caption{Review summary of large language model behavioral studies. $r$ = Pearson, $\rho$ = Spearman, $\beta$ = $\beta$-regression, $t$ = t-test, $w$ = Wilcoxon. Systematic perturbation refers to the presence of noise injected into the model or data to improve result robustness.}
        \label{tab:behav-rev}
    \end{table*}

\section{Background}

\label{sec:fan-typ-bg}
    The \textbf{\textit{fan effect}} is a psychological effect in human categorization behavior, first identified in \citet{anderson1974retrieval}, where subjects take longer to recognize and accept or reject concepts that have overlapping features with concepts previously presented in a learning set. This has most commonly been studied using concepts made up of person-place pairs. More formally, given some training concept set $S = \{<X_1,Y_1>,...,<X_n,Y_n>\}$, where $X$ and $Y$ are features of the concepts, response time when performing recognition tasks for an arbitrarily chosen query concept $<X_q, Y_q>$ is correlated with the number of times that $X_q$ and $Y_q$ occur in $S$. The effect is apparent regardless of whether or not $<X_q,Y_q>\in S$.

    Fan effects have subsequently been found to present with varying strength across different contexts. This tendency is dubbed the \textbf{\textit{differential fan effect}}. Differential fan effects have been investigated across object type and concept presentation modality. It was first identified by \citet{radvansky1991mental}, in which the fan effect was found to occur in instances where presented concepts have the same object associated with multiple places (that is to say, the object feature had a high fan value) but not when multiple persons were associated with a single place (i.e. the place feature had a high fan value). \citet{radvansky1993mental} later extended this to different object types, specifically small locations and inanimate objects. \citet{stopher1981long} found that fan effects do not seem to present when concepts are presented via images rather than text, suggesting that differential fan effect context is affected by modality in addition to content.

    There remains some debate on the mechanism of the fan effect in human subjects, particularly in regard to explaining differential fan effects. \citet{radvansky1993mental} proposed a mechanism, based on the concept of mental models, by which subjects create and maintain models of the world based on learned facts and that some types of overlap in presented concepts necessitate the creation of more models than less overlapping concept sets of the same size. \citet{anderson1999fan} proposes a different mechanism, derived from a cognitive architecture in which fan effects are mediated by changing weights of edges in the concept network. This mechanism was further supported experimentally in \citet{sohn2004differential} but challenged for larger datasets in \citet{radvansky1999fan}.

    Fan effects are found by \citet{silber1989model} in probabilistic categories created by COBWEB to be a special case of another phenomenon known as the \textbf{\textit{typicality effect}}. This would seem to suggest that fan effects may arise as a consequence of categorization, with a potential explanation being that items closer to the categorical center are more likely to collide with other items, leading to recall uncertainty, while items further from the center are less likely to experience aliasing.
    
    \textbf{\textit{Typicality}}, first formalized and identified in humans by \citet{rosch1975cognitive}, refers to a tendency of humans to perform categorization tasks quicker when prompted with a more typical member of a category than with a less typical member of a category, with level of typicality determined by how common the features of an instance of a category are among all members of the same category and among contrasting categories. That is, both an item's intra-category similarity and its inter-category similarity affect typicality assessments. For example, given pictures of two birds, a robin and a penguin, human subject response time will be higher when answering whether the penguin is a bird than whether the robin is a bird.

\subsection{Prior Work}
    \label{sec:prior-work}

    % \todo{Might help fill this out better to look for fan effects in neural models in general, not just NLP examples (also, look for NLP examples directly)}
    
    In Table \ref{tab:behav-rev}, the results of a comprehensive survey of current work in LLM cognitive behavior studies is provided. No works could be found that study language model fan effects. Though \citet{tung2024prediction} studied memory interference behavior in LLMs and use fan values in their analysis, they do not explicitly consider the fan effect or its presence. 
    
    On the other hand, work has been done that establishes the presence of typicality effects in LLMs \cite{misra2021language,bhatia2022transformer,roberts2024using} as well as vision models \cite{upadhyay2022typicality}. \citet{bhatia2022transformer} found that BERT shows evidence of typicality effects, including consistency with typicality violations common to humans. \citet{misra2021language} recreated a subset of the experiments conducted by \citet{rosch1975cognitive} which were used to identify typicality effects in humans, identifying typicality effects across numerous categories and models. \citet{roberts2024using} replicated \citet{misra2021language} with PopulationLM, establishing that the effect was not eroded when studied in a population.
    
    \citet{roberts2024using} found that the population standard deviations tended to positively correlate with typicality in encoder-only models, though not in decoder-only models. This suggests that the uncertainty captured by LLM variance may not be analogous to human uncertainty since LLMs are overwhelmingly based on the decoder-only architecture \cite{roberts2024powerful}. 

    % Typicality effects have also been shown to be present in computer vision models by \cite{upadhyay2022typicality}.
    
    %\cite{vemuri2024well} extends \cite{misra2021language, upadhyay2022typicality} to more LLMs and computer vision models, more categories, and to more recent human studies for correlation.

\section{In-Pretraining (Typicality) Fan Effect}

% Explanation of the purpose of the experiment and hypothesis

Anderson originally observed the fan effect in the response times of humans when \textbf{correctly responding} to questions. However, in \citet{silber1989model}, the authors observed human-like fan effects in a COBWEB model and found they were consistent with a special case of typicality. Based on this observation and extant work regarding the presence of typicality effects in LLMs, we hypothesize that LLMs may exhibit a fan effect induced by the relative typicality of categorical items acquired from pretraining. Specifically we formulate RQ\ref{RQ:IPF}. 

\begin{theorem2}
Given a partial list of items drawn from a category and presented to an LLM, are absence/presence prediction probabilities modulated by item typicality such that probabilities conditioned on typical items tend to be lower than those conditioned on less typical items?
\label{RQ:IPF}
\end{theorem2}

Expanding on this, based on results from \cite{roberts2024using}, more typical items tend to have increased predicted word probability even when counterfactual prompting is used, most likely due to base rate probability effects \cite{moore2024base}. However, if a fan effect is present, the probability should tend to decrease with increasing typicality. 

It is important to note that LLM probabilities are not necessarily analogous to human response times. However, existing work \cite{misra2021language,roberts2024using} has shown that typicality judgments, which have been measured via response time in humans \cite{rosch1975cognitive}, are correlated with LLM probabilities. 
\begin{figure}[h!]
    \centering
    \resizebox{\linewidth}{!}{ % Resize the picture to the line width,
    % Expert's message bubble (Outgoing)
    \begin{tcolorbox}[expertstyle]
        \hfill \textbf{In-pretraining Fan Effect Prompt} \\
        Following is a list that contains a number of birds. After the list, a bird will be judged as either present or absent in the list. If the list contains the bird, answer with present. If the list does not contain the bird, answer with absent. The list of birds is: toucan, magpie, swan, flamingo, duck, goose, blackbird, pelican, woodpecker, condor, canary, ostrich, redbird, catbird, lark, parakeet, hummingbird, bluejay, bluebird, sparrow, crow, vulture, cardinal, turkey, chicken, goldfinch, wren. According to the list, magpie is present. According to the list, kingfisher is absent. According to the list, robin is\_\_\_\_
    \end{tcolorbox}
    } % End of \resizebox

    \vspace{2mm} % Space between messages

    % LLM's message bubble (Incoming)
    \resizebox{\linewidth}{!}{ % Resize the picture to the line width,
    \begin{tcolorbox}[llmstyle]
        \textbf{LLM} \\
        $P(present)$ and $P(absent)$

    \end{tcolorbox}
    } % End of \resizebox
    \caption{Prompt to measure presence/absence belief.}
    \label{fig:Prompt1}
\vskip-0.5em

\end{figure}

\subsection{Methodology}
\label{sec:IPF-methods}

\textbf{Models:} All experimental trials are conducted among a systematically perturbed population formed from each base model using PopulationLM \cite{roberts2024using} to decrease the likelihood that obtained results are anomalous. The median value is the preferred aggregation when random sampling for the purpose of estimating a true value \cite{doerr2019resampling}. Therefore, the median across each base model population is taken as the group prediction.

We choose RoBERTa \cite{liu2019roberta}, GPT-2 \cite{radford2019language}, Llama-2 \cite{touvron2023llama}, Llama-3 \cite{meta2024introducing}, Mistral \cite{jiang2023mistral}, and SOLAR \cite{kim2023solar} as the base models for the experiments. RoBERTa and GPT-2 are chosen as representatives of models previously studied and found to exhibit typicality effects \cite{roberts2024using}. However, past work has found that higher order human-like behaviors may not be exhibited in smaller models \cite{roberts2024large}. We therefore include large open source LLMs (Llama-2, Llama-3, Mistral, and SOLAR) that may be more likely to exhibit more nuanced recall effects.

\noindent \textbf{Data Presentation: } Based on work by \citet{rosch1975cognitive} regarding human typicality judgments across items in ten categories, we construct lists for each of the ten categories in Figure \ref{fig:IPF-Details} by randomly selecting half of the items in a category. Selected items are included precisely once in a comma separated list with instructional content and two in-context examples. The in-context examples are not randomly sampled and are instead consistent across all experiments.

% \blockquote{\textit{Following is a list that contains a number of birds. After the list, a bird will be judged as either present or absent in the list. If the list contains the bird, answer with present. If the list does not contain the bird, answer with absent. The list of birds is: toucan, magpie, swan, flamingo, duck, goose, blackbird, pelican, woodpecker, condor, canary, ostrich, redbird, catbird, lark, parakeet, hummingbird, bluejay, bluebird, sparrow, crow, vulture, cardinal, turkey, chicken, goldfinch, wren. According to the list, magpie is present. According to the list, kingfisher is absent. According to the list, robin is}}

For each item (N$\approx$60) in each category and every model population member (N=50) we obtain a probability of absence and a probability of presence via counterfactual prompting \cite{moore2024base}. The probability is measured by obtaining the probability assigned to the \textit{canary} words ``present'' and ``absent'' given each constructed prompt. We repeat each experiment for each base model for each category 10 times without reuse of populations or item lists. An example interaction for the category \textit{bird} and the item \textit{robin} is shown in Figure \ref{fig:Prompt1}.

\noindent \textbf{Human Comparison:} The values for human typicality ratings are taken from \citet{rosch1975cognitive} and compared to the generated model probabilities to understand how typicality, as understood from human studies, impacts model behavior when performing recall.

\noindent \textbf{Other Hardware and Software: } All experiments used an A100 GPU Google Colab environment. Token likelihoods were obtained using a fork of the minicons Python library \cite{misra2022minicons}.

\begin{figure}[h!]
    \centering
    \includegraphics[width=\linewidth]{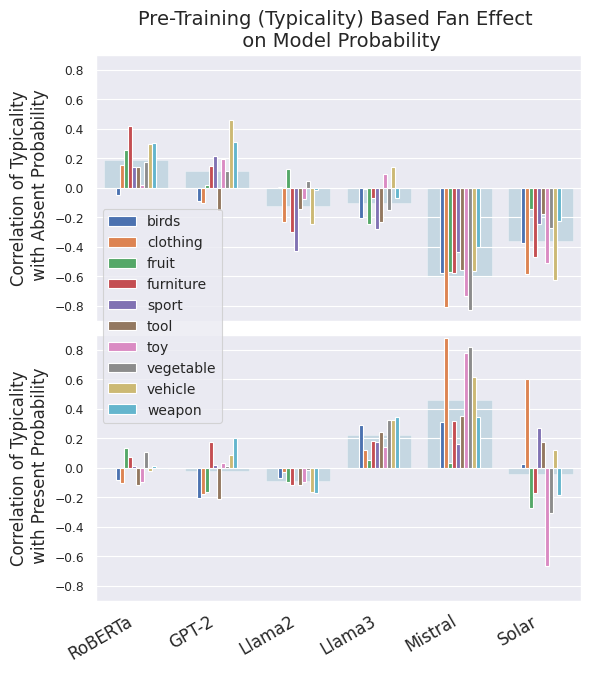} % Reduce the figure size so that it is slightly narrower than the column. Don't use precise values for figure width.This setup will avoid overfull boxes.
    \caption{\textbf{Top row:} Mistral and SOLAR show significant negative Pearson correlations consistent with fan effects across a range of categories. \textbf{Bottom row:} Items present in the context do not elicit a human-like fan effect.}
    \label{fig:IPF}
    % \vskip-1em
\end{figure}

\begin{figure*}[t!]
    \centering
    \includegraphics[width=\linewidth]{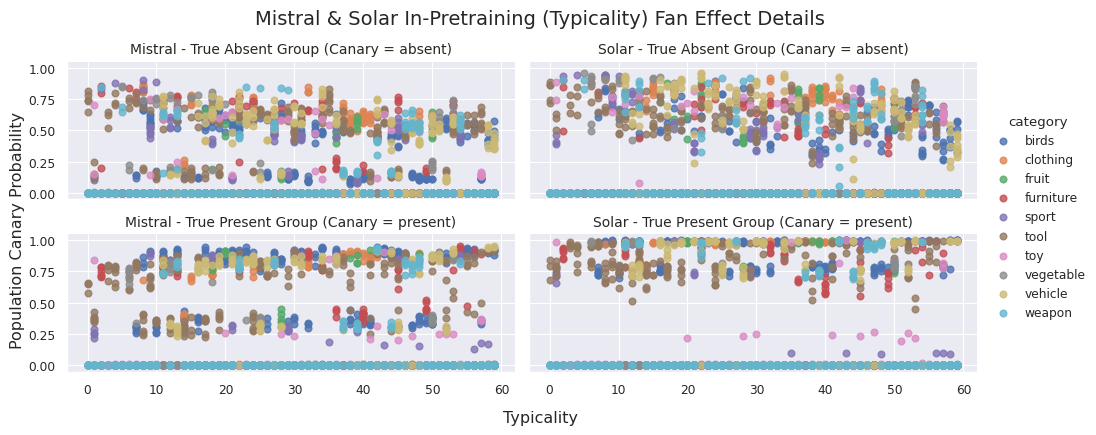} % Reduce the figure size so that it is slightly narrower than the column. Don't use precise values for figure width.This setup will avoid overfull boxes.
    \caption{\textbf{Left col:} Predictions are made using Mistral. \textbf{Right col:} Predictions are made using SOLAR. \textbf{Bottom row:} queried item is present (w/o uncertainty). \textbf{Top row:} queried item is absent (with uncertainty). Fan effects are evident in the negative Pearson correlation (shown in Figure \ref{fig:IPF}) in the natural group above the noise floor.}

    \label{fig:IPF-Details}
    % \vskip-1em
\end{figure*}

\subsection{Results}

As noted, the fan effect was only observed by Anderson in humans when responding correctly to questions. Thus, only the true absence group (TAG) and true presence group (TPG) should be considered candidate scenarios that may exhibit a human-like fan effect.

In the upper left plot in Figure \ref{fig:IPF-Details}, there is an obviously distinct group which resides above the threshold (0.35), which we refer to as the probability noise floor. We interpret the group above the noise floor to be the TAG, that is the subset of absent items which the model regards as absent. The TPG, the subset of present items which the model regards as present, can be analogously seen in the bottom left with a noise floor at (0.5). Among predictions in the TAG, the probabilities have an obvious negative correlation with typicality, showing that more typical items tend to induce lower ``absent'' probabilities. We find that SOLAR \cite{kim2023solar} shows a similar fan effect, with TAG and TPG noise floor at (0.2). 

The noise floor observed in both SOLAR and Mistral is an empirical observation which warrants additional consideration. From our investigation, the fan effect in LLMs is modulated by the probability magnitude. Therefore, low probability outputs induce noise in the observation of the fan effect in the model probabilities which are shown for completeness in Figure \ref{fig:IPF-Details} but filtered in the correlation analysis shown in Figure \ref{fig:IPF}.

Interestingly, in the lower left of Figure \ref{fig:IPF-Details} the TPG for Mistral has positive correlations which are inconsistent with the fan effect. This is reflected in the bottom of Figure \ref{fig:IPF} as well. SOLAR, on the other hand, tends toward inter-category randomness in the bottom of Figure \ref{fig:IPF}.

\subsection{Discussion}

In response to RQ\ref{RQ:IPF}, we find in Figure \ref{fig:IPF-Details} that items absent from the list elicit a human-consistent fan effect evident in the canary probabilities in Mistral \cite{jiang2023mistral} and SOLAR \cite{kim2023solar}. The probabilities show a significant (r>0.3) \cite{hinkle2003applied} correlation with intra-category typicality in Figure \ref{fig:IPF} consistent with the fan effects discovered in COBWEB and theorized in humans. This result shows that LLMs exhibit fan effects based on the effects of typicality present in the pretraining data.  

RoBERTa \cite{liu2019roberta}, GPT-2 \cite{radford2019language}, Llama-2 \cite{touvron2023llama}, and Llama-3 \cite{meta2024introducing} were equivalently evaluated but showed no significant correlation, though Llama-3 does show a similar, slight effect. We additionally conducted the correlation investigation presented using the population variance in place of the token probabilities and found no significant correlations. This reinforces the possibility put forth in \citet{roberts2024using} that decoder-only LLM variance may not capture human-like uncertainty given fan effects are understood as an expression of human uncertainty. 

\noindent \textbf{Interpretation:} We were surprised to find the fan effect exhibited in the TAG but not the TPG. However, in retrospect this could have been anticipated based on nuanced consideration of the experiment.  

The fan effect is canonically explained as a modulation of human uncertainty based on the categorical distance from an exemplar. When evaluating the TPG, the model is able to judge with near certainty by retrieving the queried item. On the other hand when judging the absence of a TAG item, the model can only know that the item has not been retrieved. The model assigns the probability of absence although it may actually be that the item is present but overlooked, inducing uncertainty. We hypothesize this uncertainty is precisely what the fan effect is modulating. So, when queried about an absent atypical item, the model responds confidently as if implying, ``I definitely didn't see \textit{that}''.

The above scenario in which the fan effect is only observed in the absent case seems plausibly consistent with human cognitive behavior. Imagine a context in which a human has a deck of cards and is asked if a card is present. If the card is found, then the person will have no uncertainty about their response. On the other hand, if the card is not found, the certainty of the response would be expected to be modulated by the fan effect. That is, if an unusual or outlier card is being searched for then it is likely that the person would notice if it had been present. However, it is reasonable that a human could more easily overlook a common card.

We hypothesize that the uncertainty mitigation due to access to the queried items in the TPG leads to the disruption of the fan effect in Mistral and SOLAR. Our results leave unclear the nature of the fan effect under mitigated uncertainty in the TPG.

\subsection{Next Steps}

Future work should consider creating long context lists that prevent models from retrieving TPG items with high fidelity to attempt to induce uncertainty and fan effects in the TPG. This was not possible currently since no extant lists of intra-category typical items in humans are sufficiently long. However, it may be possible to use LLMs to augment the typicality datasets to create a sufficiently large list.

Results from Mistral suggest that fan effects without uncertainty tend toward a typicality effect response with increasing probability as typicality increases. However, results from SOLAR suggest that they tend toward noise. Future work should additionally attempt to disambiguate the nature of the fan effect when uncertainty is mitigated. 

Future work should investigate human behavior in a scenario similar to the described card experiment to understand human fan effect behavior under mitigated uncertainty.

\section{In-Context Fan Effect}
We investigate the presence of fan effects as originally defined in \citet{anderson1974retrieval} in the context of concepts composed of categorical features. This addresses the question of whether fan effects show up in concepts defined and fan values induced exclusively in-context. We formulate this as RQ\ref{RQ:ICF}. We augment our analysis to investigate the presence of differential fan effect as described in \citet{radvansky1991mental}, providing RQ\ref{RQ:ICF-DFE}.

\begin{theorem2}
    Given a list of simple concepts defined by their composite features that is presented to an LLM, are absence/presence prediction probabilities modulated by feature fan values such that probabilities conditioned on high fan features tend to be lower than probabilities conditioned on low fan features?
    \label{RQ:ICF}
\end{theorem2}

\begin{theorem2}
    Given a list of simple concepts defined by their composite features that is presented to an LLM, is correlation of absence/presence prediction probability with fan value modulated by the fan values of one feature more strongly than another feature?
    \label{RQ:ICF-DFE}
\end{theorem2}

% \begin{theorem2}
% Given a partial list of items drawn from a category and presented to an LLM, are absence/presence prediction probabilities modulated by item typicality such that probabilities conditioned on typical items tend to be lower than those conditioned on less typical items?
% \label{RQ:IPF}
% \end{theorem2}

\subsection{Methodology}

We closely recreate the experimental methodology of \citet{anderson1974retrieval}, with methods similar to those described in section \ref{sec:IPF-methods} for in-pretraining fan effects.

\textbf{Models: } Based on the results regarding in-pretraining fan effects, we conduct in-context fan effect experiments with populations formed from Mistral and SOLAR using PopulationLM. The experiment uses a generated model population of size $N=50$ with median aggregation across population to determine group prediction. As before, probabilities are obtained using the \textit{canary} words ``present'' and ``absent''.

\noindent \textbf{Data Presentation: }  Concepts are defined as natural language facts that pair persons, in the form of occupation labels, with places. Each fact is presented as a sentence of the form ``The <occupation> is in the <place>''. Features are sampled from predefined person and place lists, each of size 20. The fan value is defined as the number of concepts that contain a given feature value. For example, if three distinct concepts indicate a person is present in the place ``School'', the fan value of ``School'' is 3. Concept lists are randomly generated to control for ordering effects and feature combination base rates due to semantically connected features (e.g. <Priest, Church>). 

\begin{table}[h]
  \centering
    \begin{tabular}{c c p{1.2cm} p{1.2cm} p{1.2cm}}
\cmidrule{3-5}
    \multicolumn{1}{c}{}                                          &                       & \multicolumn{3}{c}{No. of Concepts per Person} \\
\cmidrule{3-5}
    \multicolumn{1}{c}{}                                          &                       & 1     & 2     & 3 \\
    \midrule
    \multirow{9}[6]{*}{\rotatebox{90}{No. of Concepts per Place}} & \multirow{3}[2]{*}{1} & aA    & dD    & gG \\
                                                                  &                       & bB    & eE    & hH \\
                                                                  &                       & cC    & fF    & iI \\
\cmidrule{2-5}                                                    & \multirow{3}[2]{*}{2} & jJ    & eK    & gJ \\
                                                                  &                       & kK    & rR    & hR \\
                                                                  &                       & lL    &       & iL \\
\cmidrule{2-5}                                                    & \multirow{3}[2]{*}{3} & mM    & dM    & gM \\
                                                                  &                       & nN    & rN    & hN \\
                                                                  &                       & oO    & fO    & iO \\
    \bottomrule
    \end{tabular}%
  \caption{Feature assignment pattern used in \citet{anderson1999fan} and replicated in the in-context fan effect experiment.}%
  \label{tab:AndersonTruth}%
\end{table}%

The concepts in the recreation of Anderson are generated exactly as in \citet{anderson1974retrieval}. A predefined set of feature combinations are used, as summarized in Table \ref{tab:AndersonTruth}, which are designated by lowercase letters for persons and uppercase letters for places. The person and place assigned to each letter is randomly selected without replacement at the beginning of each trial. The result is N=26 concepts presented to the model in each trial, with a total of 16 fan value combinations (including fan = 0 for features not present in the set).

% The Anderson Augmented experiment randomly generates every concept independently. Concepts are generated, disallowing for duplicates, until the set contains N=50 concepts. Each concept is formed by selecting a person and place with replacement via Gaussian sampling. Gaussian sampling is used to facilitate a diverse range of feature fan combinations. Random sampling allows for a larger range of feature fan values, with a maximum possible fan of 20 (when a person is associated with every place or vice versa).

Prompts presented to the model follow prompt design similar to that in section \ref{sec:IPF-methods}. The prompt is composed of four sections: An instructional preamble, the concept list, a two-shot ICL example, and the test query. The ICL examples include a concept that is appended to the end of the concept list that is guaranteed to not be generated. This guaranteed concept is followed by two example queries and simulated outputs, one where the concept is the guaranteed present concept and one with a guaranteed absent concept. 

An example prompt in which the concept <Doctor, Park> is shown in Figure \ref{fig:Prompt2}. Note that <Mechanic, Mall> is included in all trials and has a guaranteed fan value of 1 for both features, while <Airport, Pilot> is absent in all trials.

\begin{figure}[h!]
    \centering
    \resizebox{\linewidth}{!}{ % Resize the picture to the line width,
    % Expert's message bubble (Outgoing)
    \begin{tcolorbox}[expertstyle]
        \hfill \textbf{In-Context Fan Effect Prompt} \\
        Following is a list that contains a number of people and the places in which they are located. After the list, a person will be judged as either present or absent in a specified place. When asked about person A in place B, if the list says that person A is in place B, answer with present. If the list does not say that person A is in place B, answer with absent. The list of people and places is: The Nurse is in the Studio. The Police Officer is in the Bank. \ldots The Mechanic is in the Mall. According to the list, in the Mall, the Mechanic is present. According to the list, in the Airport, the Pilot is absent. According to the list, in the Park, the Doctor is\_\_\_\_
    \end{tcolorbox}
    } % End of \resizebox

    \vspace{2mm} % Space between messages

    % LLM's message bubble (Incoming)
    \resizebox{\linewidth}{!}{ % Resize the picture to the line width,
    \begin{tcolorbox}[llmstyle]
        \textbf{LLM} \\
        $P(present)$ and $P(absent)$

    \end{tcolorbox}
    } % End of \resizebox
    \caption{Prompt to measure presence/absence belief.}
    \label{fig:Prompt2}
\vskip-0.5em

\end{figure}

% \blockquote{
%     \textit{Following is a list that contains a number of people and the places in which they are located. After the list, a person will be judged as either present or absent in a specified place. When asked about person A in place B, if the list says that person A is in place B, answer with present. If the list does not say that person A is in place B, answer with absent. The list of people and places is: The Nurse is in the Studio. The Police Officer is in the Bank. \ldots The Mechanic is in the Mall. According to the list, in the Mall, the Mechanic is present. According to the list, in the Airport, the Pilot is absent. According to the list, in the Park, the Doctor is}
% }

\noindent \textbf{Human Comparison: } The data pairings generated are based on the data presented to humans in \citet{anderson1999fan} which were shown to illicit the fan effect in human recall. 

\noindent \textbf{Other Hardware and Software: } All experiments are conducted on an A100 GPU Google Colab environment. Token likelihoods were again obtained with a modified version of the minicons library \cite{misra2022minicons}.

\subsection{Results}

\begin{figure}[h]
    \centering
    \includegraphics[width=\linewidth]{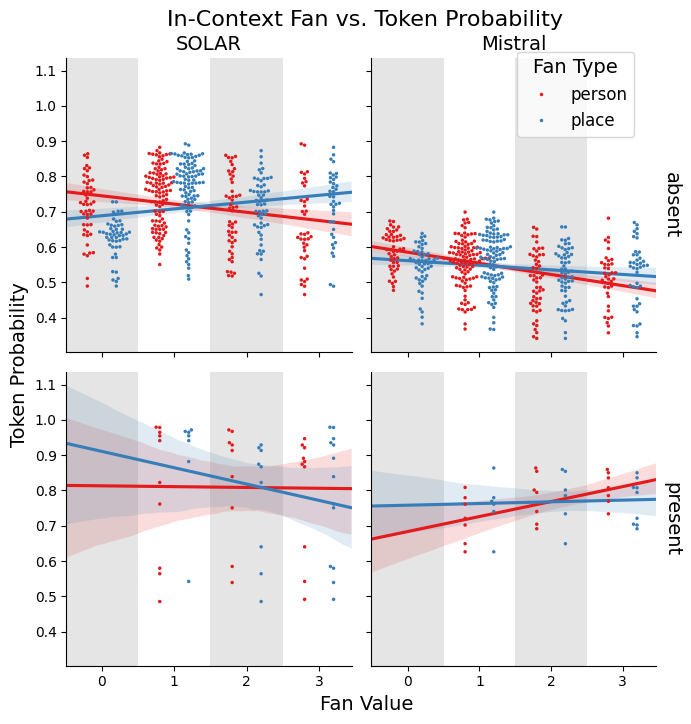} % Reduce the figure size so that it is slightly narrower than the column. Don't use precise values for figure width.This setup will avoid overfull boxes.
    \caption{Results of the Anderson recreation experiments on SOLAR and Mistral \textbf{Top row:} queried item is absent with the model predicting true absence (with uncertainty). \textbf{Bottom row:} queried item is present with the model predicting true presence (w/o uncertainty). Lines of best fit are included. Pearson correlations shown in Figure \ref{fig:ICF-Details}.}
    \label{fig:ICF}
    % \vskip-1em
\end{figure}

\begin{figure}[h]
    \centering
    \includegraphics[width=\linewidth]{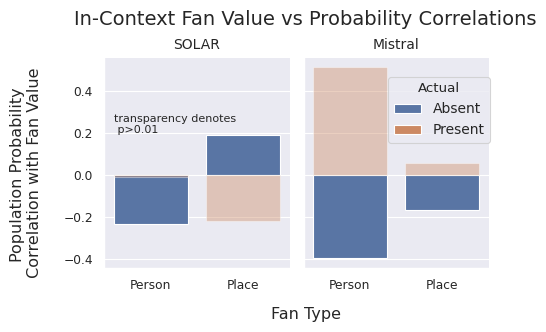} % Reduce the figure size so that it is slightly narrower than the column. Don't use precise values for figure width.This setup will avoid overfull boxes.
    \caption{Negative correlations when the queried item is absent suggests items are recalled with higher certainty when the item has fewer in-context appearances (low fan value). Fan values derived from the queried person show fan effects while place fan values cause a distruption of the fan effect. No ``present'' item queries have significant p values though all ``absent'' item queries do. }
    \label{fig:ICF-Details}
    % \vskip-1em
\end{figure}

The results for both models are shown in Figure \ref{fig:ICF}. As was the case in the in-pretraining experiments, a probability noise floor was noted in the data for both canary completions (Mistral-absent: 0.3; Mistral-present: 0.4; SOLAR-absent: 0.45; SOLAR-present: 0.4), providing a TAG and TPG. The figures are truncated to show only the TPG and TAG datapoints. Correlation statistics of the results are shown in Figure \ref{fig:ICF-Details}, with solid columns indicating correlations with a $p\leq0.01$.

In Mistral, we once again see an obvious negative correlation between canary probability and fan value in the TAG predictions. This is consistent with a fan effect when evaluating absence of a concept (RQ \ref{RQ:ICF}). In the TAG, we see a stronger correlation with the fan value of the person feature than with the fan value of the place feature, supporting a positive result for RQ \ref{RQ:ICF-DFE}. This is consistent with results regarding differential fan effects in \citet{radvansky1991mental}, which found that the fan effect is mediated more by the fan of a particular object than the fan of a particular location. 

SOLAR shows a slightly different story. For the TAG predictions, we still see a significant negative correlation when correlating with the fan of person, but a positive correlation with fan of place. TPG predictions instead show a negative correlation against fan of place and no correlation against fan of person. While this seems inconsistent with our Mistral results, it is consistent with our prior interpretations when properly analyzed. Based on these results, SOLAR and Mistral both show evidence of the fan effect in, at minimum, the same situations as in humans, which is to say uncertain contexts and based on the fan of person.

From the in-pretraining experiment, we expect that mitigated uncertainty in the TPG may lead to disruption of the fan effect. In confirmation, among TPG items all correlations fail to achieve a significant p value for fan value and canary probability Pearson correlation, again suggesting that mitigated uncertainty disrupts the fan effect.

% The results of both experiments are shown in Figure \ref{fig:ICF}. The canary probabilities are analyzed for cases where the canary matches the true category of the concept. To investigate differential fan effects, the resulting canary probabilities are graphed according to the fan value of each feature independently. All results showed a natural splitting point and the probabilities below that point are omitted from the graph. As shown, there is a moderate negative Spearman's-r correlation between fan value and token probability for both experiments in the absent case, consistent with the fan effect. As in human studies of differential fan effect, we see a stronger correlation for person fan than place fan. Interestingly, in the present case, there is a comparable positive correlation, suggesting an inverse fan effect occurs when the task has no uncertainty.

\subsection{Next Steps}
There are numerous enhancements that could be applied to these experiments. While occupations were chosen as proxies for persons to be consistent with \citet{anderson1974retrieval}, more unique identifiers like names may yield a stronger differential fan effect if the mental models mechanism proposed by \cite{radvansky1991mental} is present in language models. This should be tested empirically in future work to investigate the nature of differential fan effects. Additionally, other feature types that are not related to persons and places should be investigated. 

% Negative concepts may also allow exploring our hypothesized explanation for our results by reversing the uncertainty relationship between presence and absence. 

Human cognitive experiments often include a dimension of elapsed time between training and testing time when studying memory-sensitive behaviors. Future work should consider simulating this time separation in language models. Though language models do not possess a directly analogous temporal dimension, experiments could evaluate the injection of semantic noise of varying length as a potential proxy. In fact such an experiment may suggest that time, to humans, is itself a form of semantic noise. 

% Explanation of the purpose of the experiments and hypothesis

%Consistent with nuanced experiments regarding the fan effect. The effect is more pronounced in fan of the person as a mental model that permits a collocated person is less intuitive than a model of a place that permits multiple people at that location. 

\section{Conclusions}

Our experiments are the first to evaluate LLMs for the presence of human-like fan effects. We have shown that Mistral and SOLAR have learned to exhibit fan effects from training on human language data. This paper is not the first to identify SOLAR and Mistral as important human-like LLMs. \citet{roberts2024large} found SOLAR and Mistral to be significantly more human-like than a large body of other open-source models when evaluated in a game theoretic context. Given Mistral was built from Llama-2 and SOLAR was built from Mistral, the authors propose the more human-like behavior may be the result of an improved representation acquired through additional training of Mistral with sliding window attention. 

Our results show that fan effects are present both when the fan value is induced in-pretraining in the form of intra-category typicality and when the fan value is induced in-context in the form of repeated items within a list. The presence of typicality-based fan effects in language models lends further credence to the findings of \citet{silber1989model} suggesting that fan effects are a special case of typicality effects.

Additionally, we find that when uncertainty is mitigated, the fan effect is disrupted with divergent disruption patterns across LLMs. The divergent patterns across Mistral and SOLAR beg further investigation. However, we are unaware of any cognitive science literature that addresses fan effects in a disruptive scenario with mitigated uncertainty. Therefore, it is unclear how a human may behave in a similar context. We therefore call for human experiments. 

Similarly, when the fan value is derived from place instead of person in the Anderson experiment, both Mistral and SOLAR exhibit a disruption of the fan effect in agreement with nuanced work regarding differential fan effects \cite{radvansky1991mental}. Again, each of these models diverges in the nature of the disruption but shows a consistent pattern of fan effects in the case of true absence when the fan value is calculated on the person feature. 

Finally, we hope this paper will prove synergistic with the wider cognitive science and computational linguistic communities. By adapting experiments to evaluate the presence of known human cognitive effects in LLMs, we may gain new insight into cognitive effects. These insights not only help to explain the factors which influence the behavior of complex language models but also provide new potential hypotheses regarding the cognitive behavior of humans.

\section{Practical Implications}

Human-like uncertainty is shown to be present in Mistral and SOLAR in the form of a fan effect both when the fan value is induced in the pretraining of the model and in the context. However, just as found in \citet{roberts2024using}, the common measures of model uncertainty, variance and standard deviation, may not tend to correlate well with human uncertainty as quantified by the fan effect. This suggests that more work needs to be done to develop a human-consistent measure of LLM uncertainty. 

Additionally, the fan effect should be considered when engaging LLMs in applications that require recall. The results here suggest that LLMs may have more trouble correctly evaluating the presence or absence of (1) items when the item is frequently present in the pretraining data and (2) coincident items when the base item is frequently present in the context of the model.

\section*{Acknowledgments}

We appreciate the helpful recommendations of reviewers that led to improved presentation clarity and the insightful evaluation of the area chair.

\section*{Limitations}

While this paper demonstrates that LLMs exhibit fan effects. It may be the case that the observed effects tend to be weak in comparison to the probability magnitude. So, it is unclear if LLMs fail in recall scenarios in manners consistent with fan effects. 

% \section{Ethical Considerations}

% Doesn't count toward page limit

\bibliography{anthology,custom}

\begin{thebibliography}{45}
\providecommand{\natexlab}[1]{#1}

\bibitem[{Anderson and Reder(1999)}]{anderson1999fan}
John~R Anderson and Lynne~M Reder. 1999.
\newblock The fan effect: New results and new theories.
\newblock \emph{Journal of Experimental Psychology: General}, 128(2):186.

\bibitem[{Anderson(1974)}]{anderson1974retrieval}
John~Robert Anderson. 1974.
\newblock Retrieval of propositional information from long-term memory.
\newblock \emph{Cognitive psychology}, 6(4):451--474.

\bibitem[{Bhatia and Richie(2022)}]{bhatia2022transformer}
Sudeep Bhatia and Russell Richie. 2022.
\newblock Transformer networks of human conceptual knowledge.
\newblock \emph{Psychological Review}.

\bibitem[{Binz and Schulz(2023)}]{binz2023using}
Marcel Binz and Eric Schulz. 2023.
\newblock Using cognitive psychology to understand gpt-3.
\newblock \emph{Proceedings of the National Academy of Sciences}, 120(6):e2218523120.

\bibitem[{Bubeck et~al.(2023)Bubeck, Chandrasekaran, Eldan, Gehrke, Horvitz, Kamar, Lee, Lee, Li, Lundberg et~al.}]{bubeck2023sparks}
S{\'e}bastien Bubeck, Varun Chandrasekaran, Ronen Eldan, Johannes Gehrke, Eric Horvitz, Ece Kamar, Peter Lee, Yin~Tat Lee, Yuanzhi Li, Scott Lundberg, et~al. 2023.
\newblock Sparks of artificial general intelligence: Early experiments with gpt-4.
\newblock \emph{arXiv preprint arXiv:2303.12712}.

\bibitem[{Coda-Forno et~al.(2024)Coda-Forno, Binz, Wang, and Schulz}]{coda2024cogbench}
Julian Coda-Forno, Marcel Binz, Jane~X Wang, and Eric Schulz. 2024.
\newblock \href {https://openreview.net/forum?id=Q3104y8djk} {Cogbench: a large language model walks into a psychology lab}.
\newblock In \emph{Forty-first International Conference on Machine Learning}.

\bibitem[{Coda-Forno et~al.(2023)Coda-Forno, Witte, Jagadish, Binz, Akata, and Schulz}]{coda2023inducing}
Julian Coda-Forno, Kristin Witte, Akshay~K Jagadish, Marcel Binz, Zeynep Akata, and Eric Schulz. 2023.
\newblock Inducing anxiety in large language models increases exploration and bias.
\newblock \emph{arXiv preprint arXiv:2304.11111}.

\bibitem[{Doerr and Sutton(2019)}]{doerr2019resampling}
Benjamin Doerr and Andrew~M Sutton. 2019.
\newblock When resampling to cope with noise, use median, not mean.
\newblock In \emph{Proceedings of the Genetic and Evolutionary Computation Conference}, pages 242--248.

\bibitem[{Hagendorff et~al.(2023)Hagendorff, Fabi, and Kosinski}]{hagendorff2023human}
Thilo Hagendorff, Sarah Fabi, and Michal Kosinski. 2023.
\newblock Human-like intuitive behavior and reasoning biases emerged in large language models but disappeared in chatgpt.
\newblock \emph{Nature Computational Science}, 3(10):833--838.

\bibitem[{Hinkle et~al.(2003)Hinkle, Wiersma, Jurs et~al.}]{hinkle2003applied}
Dennis~E Hinkle, William Wiersma, Stephen~G Jurs, et~al. 2003.
\newblock \emph{Applied statistics for the behavioral sciences}, volume 663.
\newblock Houghton Mifflin Boston.

\bibitem[{Jiang et~al.(2023)Jiang, Sablayrolles, Mensch, Bamford, Chaplot, Casas, Bressand, Lengyel, Lample, Saulnier et~al.}]{jiang2023mistral}
Albert~Q Jiang, Alexandre Sablayrolles, Arthur Mensch, Chris Bamford, Devendra~Singh Chaplot, Diego de~las Casas, Florian Bressand, Gianna Lengyel, Guillaume Lample, Lucile Saulnier, et~al. 2023.
\newblock Mistral 7b.
\newblock \emph{arXiv preprint arXiv:2310.06825}.

\bibitem[{Jones and Steinhardt(2022)}]{jones2022capturing}
Erik Jones and Jacob Steinhardt. 2022.
\newblock Capturing failures of large language models via human cognitive biases.
\newblock \emph{Advances in Neural Information Processing Systems}, 35:11785--11799.

\bibitem[{Kim et~al.(2023)Kim, Park, Kim, Lee, Song, Kim, Kim, Kim, Lee, Kim et~al.}]{kim2023solar}
Dahyun Kim, Chanjun Park, Sanghoon Kim, Wonsung Lee, Wonho Song, Yunsu Kim, Hyeonwoo Kim, Yungi Kim, Hyeonju Lee, Jihoo Kim, et~al. 2023.
\newblock Solar 10.7 b: Scaling large language models with simple yet effective depth up-scaling.
\newblock \emph{arXiv preprint arXiv:2312.15166}.

\bibitem[{Kosinski(2023)}]{kosinski2023theory}
Michal Kosinski. 2023.
\newblock Theory of mind may have spontaneously emerged in large language models.
\newblock \emph{arXiv preprint arXiv:2302.02083}.

\bibitem[{Lampinen et~al.(2023)Lampinen, Dasgupta, Chan, Sheahan, Creswell, Kumaran, McClelland, and Hill}]{dasgupta2023language}
Andrew~K Lampinen, Ishita Dasgupta, Stephanie~CY Chan, Hannah~R Sheahan, Antonia Creswell, Dharshan Kumaran, James~L McClelland, and Felix Hill. 2023.
\newblock Language models show human-like content effects on reasoning tasks.
\newblock \emph{arXiv preprint arXiv:2207.07051}.

\bibitem[{Lamprinidis(2023)}]{lamprinidis2023llm}
Sotiris Lamprinidis. 2023.
\newblock Llm cognitive judgements differ from human.
\newblock \emph{arXiv preprint arXiv:2307.11787}.

\bibitem[{Li et~al.(2023)Li, Chong, Stepputtis, Campbell, Hughes, Lewis, and Sycara}]{li2023theory}
Huao Li, Yu~Quan Chong, Simon Stepputtis, Joseph Campbell, Dana Hughes, Michael Lewis, and Katia Sycara. 2023.
\newblock Theory of mind for multi-agent collaboration via large language models.
\newblock \emph{arXiv preprint arXiv:2310.10701}.

\bibitem[{Liu et~al.(2019)Liu, Ott, Goyal, Du, Joshi, Chen, Levy, Lewis, Zettlemoyer, and Stoyanov}]{liu2019roberta}
Yinhan Liu, Myle Ott, Naman Goyal, Jingfei Du, Mandar Joshi, Danqi Chen, Omer Levy, Mike Lewis, Luke Zettlemoyer, and Veselin Stoyanov. 2019.
\newblock Roberta: A robustly optimized bert pretraining approach.
\newblock \emph{arXiv preprint arXiv:1907.11692}.

\bibitem[{Ma et~al.(2023)Ma, Sansom, Peng, and Chai}]{ma2023towards}
Ziqiao Ma, Jacob Sansom, Run Peng, and Joyce Chai. 2023.
\newblock Towards a holistic landscape of situated theory of mind in large language models.
\newblock \emph{arXiv preprint arXiv:2310.19619}.

\bibitem[{McCoy et~al.(2019)McCoy, Pavlick, and Linzen}]{mccoy2019right}
R~Thomas McCoy, Ellie Pavlick, and Tal Linzen. 2019.
\newblock Right for the wrong reasons: Diagnosing syntactic heuristics in natural language inference.
\newblock \emph{arXiv preprint arXiv:1902.01007}.

\bibitem[{Meta(2024)}]{meta2024introducing}
AI~Meta. 2024.
\newblock Introducing meta llama 3: The most capable openly available llm to date.
\newblock \emph{Meta AI.}

\bibitem[{Michaelov et~al.(2023)Michaelov, Arnett, Chang, and Bergen}]{michaelov-etal-2023-structural}
James Michaelov, Catherine Arnett, Tyler Chang, and Ben Bergen. 2023.
\newblock \href {https://doi.org/10.18653/v1/2023.emnlp-main.227} {Structural priming demonstrates abstract grammatical representations in multilingual language models}.
\newblock In \emph{Proceedings of the 2023 Conference on Empirical Methods in Natural Language Processing}, pages 3703--3720, Singapore. Association for Computational Linguistics.

\bibitem[{Misra(2022)}]{misra2022minicons}
Kanishka Misra. 2022.
\newblock minicons: Enabling flexible behavioral and representational analyses of transformer language models.
\newblock \emph{arXiv preprint arXiv:2203.13112}.

\bibitem[{Misra et~al.(2021)Misra, Ettinger, and Rayz}]{misra2021language}
Kanishka Misra, Allyson Ettinger, and Julia~Taylor Rayz. 2021.
\newblock Do language models learn typicality judgments from text?
\newblock \emph{arXiv preprint arXiv:2105.02987}.

\bibitem[{Moore et~al.(2024)Moore, Roberts, Pham, Ewaleifoh, and Fisher}]{moore2024base}
Kyle Moore, Jesse Roberts, Thao Pham, Oseremhen Ewaleifoh, and Doug Fisher. 2024.
\newblock The base-rate effect on llm benchmark performance: Disambiguating test-taking strategies from benchmark performance.
\newblock \emph{arXiv preprint arXiv:2406.11634}.

\bibitem[{Radford et~al.(2019)Radford, Wu, Child, Luan, Amodei, Sutskever et~al.}]{radford2019language}
Alec Radford, Jeffrey Wu, Rewon Child, David Luan, Dario Amodei, Ilya Sutskever, et~al. 2019.
\newblock Language models are unsupervised multitask learners.
\newblock \emph{OpenAI blog}, 1(8):9.

\bibitem[{Radvansky and Zacks(1991)}]{radvansky1991mental}
GA~Radvansky and RT~Zacks. 1991.
\newblock Mental models and fact retrieval.
\newblock \emph{Journal of Experimental Psychology: Learning, Memory, and Cognition}, 17:940--953.

\bibitem[{Radvansky(1999)}]{radvansky1999fan}
Gabriel~A Radvansky. 1999.
\newblock The fan effect: a tale of two theories.

\bibitem[{Radvansky et~al.(1993)Radvansky, Spieler, and Zacks}]{radvansky1993mental}
Gabriel~A Radvansky, Daniel~H Spieler, and Rose~T Zacks. 1993.
\newblock Mental model organization.
\newblock \emph{Journal of Experimental Psychology: Learning, Memory, and Cognition}, 19(1):95.

\bibitem[{Roberts(2024)}]{roberts2024powerful}
Jesse Roberts. 2024.
\newblock How powerful are decoder-only transformer neural models?
\newblock In \emph{2024 International Joint Conference on Neural Networks (IJCNN)}, pages 1--8. IEEE.

\bibitem[{Roberts et~al.(2024{\natexlab{a}})Roberts, Moore, and Fisher}]{roberts2024large}
Jesse Roberts, Kyle Moore, and Doug Fisher. 2024{\natexlab{a}}.
\newblock Do large language models learn human-like strategic preferences?
\newblock \emph{arXiv preprint arXiv:2404.08710}.

\bibitem[{Roberts et~al.(2024{\natexlab{b}})Roberts, Moore, Wilenzick, and Fisher}]{roberts2024using}
Jesse Roberts, Kyle Moore, Drew Wilenzick, and Douglas Fisher. 2024{\natexlab{b}}.
\newblock \href {https://doi.org/10.1609/aaai.v38i17.29856} {Using artificial populations to study psychological phenomena in neural models}.
\newblock \emph{Proceedings of the AAAI Conference on Artificial Intelligence}, 38(17):18906--18914.

\bibitem[{Rosch(1975)}]{rosch1975cognitive}
Eleanor Rosch. 1975.
\newblock Cognitive representations of semantic categories.
\newblock \emph{Journal of experimental psychology: General}, 104(3):192.

\bibitem[{Sap et~al.(2022)Sap, LeBras, Fried, and Choi}]{sap2022neural}
Maarten Sap, Ronan LeBras, Daniel Fried, and Yejin Choi. 2022.
\newblock Neural theory-of-mind? on the limits of social intelligence in large lms.
\newblock \emph{arXiv preprint arXiv:2210.13312}.

\bibitem[{Silber and Fisher(1989)}]{silber1989model}
J~Silber and DH~Fisher. 1989.
\newblock A model of natural category structure and its behavioral implications.
\newblock In \emph{Proceedings of the eleventh annual conference of the Cognitive Science Society}, pages 884--891. Erlbaum Hillsdale, NJ.

\bibitem[{Sinclair et~al.(2022)Sinclair, Jumelet, Zuidema, and Fern{\'a}ndez}]{sinclair2022structural}
Arabella Sinclair, Jaap Jumelet, Willem Zuidema, and Raquel Fern{\'a}ndez. 2022.
\newblock Structural persistence in language models: Priming as a window into abstract language representations.
\newblock \emph{Transactions of the Association for Computational Linguistics}, 10:1031--1050.

\bibitem[{Sohn et~al.(2004)Sohn, Anderson, Reder, and Goode}]{sohn2004differential}
Myeong-Ho Sohn, John~R Anderson, Lynne~M Reder, and Adam Goode. 2004.
\newblock Differential fan effect and attentional focus.
\newblock \emph{Psychonomic Bulletin \& Review}, 11:729--734.

\bibitem[{Stopher and Kirsner(1981)}]{stopher1981long}
Kerry Stopher and Kim Kirsner. 1981.
\newblock Long-term memory for pictures and sentences.
\newblock \emph{Memory \& Cognition}, 9:34--40.

\bibitem[{Suri et~al.(2023)Suri, Slater, Ziaee, and Nguyen}]{suri2023large}
Gaurav Suri, Lily~R Slater, Ali Ziaee, and Morgan Nguyen. 2023.
\newblock Do large language models show decision heuristics similar to humans? a case study using gpt-3.5.
\newblock \emph{arXiv preprint arXiv:2305.04400}.

\bibitem[{Touvron et~al.(2023)Touvron, Martin, Stone, Albert, Almahairi, Babaei, Bashlykov, Batra, Bhargava, Bhosale et~al.}]{touvron2023llama}
Hugo Touvron, Louis Martin, Kevin Stone, Peter Albert, Amjad Almahairi, Yasmine Babaei, Nikolay Bashlykov, Soumya Batra, Prajjwal Bhargava, Shruti Bhosale, et~al. 2023.
\newblock Llama 2: Open foundation and fine-tuned chat models.
\newblock \emph{arXiv preprint arXiv:2307.09288}.

\bibitem[{Trott et~al.(2023)Trott, Jones, Chang, Michaelov, and Bergen}]{trott2023large}
Sean Trott, Cameron Jones, Tyler Chang, James Michaelov, and Benjamin Bergen. 2023.
\newblock Do large language models know what humans know?
\newblock \emph{Cognitive Science}, 47(7):e13309.

\bibitem[{Tung(2024)}]{tung2024prediction}
Tzu-Yun Tung. 2024.
\newblock \emph{Prediction and Memory Retrieval in Dependency Resolution}.
\newblock Ph.D. thesis.

\bibitem[{Ullman(2023)}]{ullman2023large}
Tomer Ullman. 2023.
\newblock Large language models fail on trivial alterations to theory-of-mind tasks.
\newblock \emph{arXiv preprint arXiv:2302.08399}.

\bibitem[{Upadhyay et~al.(2022)Upadhyay, Mittal, and Varma}]{upadhyay2022typicality}
Neha Upadhyay, Kritika Mittal, and Sashank Varma. 2022.
\newblock Typicality gradients in computer vision models.
\newblock In \emph{Proceedings of the Annual Meeting of the Cognitive Science Society}, volume~44.

\bibitem[{Yax et~al.(2024)Yax, Anll{\'o}, and Palminteri}]{yax2024studying}
Nicolas Yax, Hernan Anll{\'o}, and Stefano Palminteri. 2024.
\newblock Studying and improving reasoning in humans and machines.
\newblock \emph{Communications Psychology}, 2(1):51.

\end{thebibliography}

% \bibliography{custom}

\end{document}